\pdfoutput=1

\documentclass[11pt]{article}

\usepackage{acl}

\usepackage{times}
\usepackage{latexsym}

\usepackage{CJKutf8}
\usepackage{booktabs}
\usepackage{graphicx}
\usepackage{subfigure}
\usepackage{comment}
\usepackage{xspace}
\usepackage{adjustbox}
\usepackage{mathtools}
\usepackage{bbm}
\usepackage{multirow}
\usepackage{amssymb}
\usepackage{bbding}
\usepackage{amsmath}
\usepackage{verbatim}
\usepackage{amsfonts}
\usepackage{bm}
\usepackage[normalem]{ulem}

\usepackage[T1]{fontenc}

\usepackage[utf8]{inputenc}

\usepackage{microtype}
\usepackage{xcolor}
\usepackage{fancyhdr}

\usepackage{inconsolata}

\usepackage{hyperref}
\usepackage{url}
\usepackage{graphicx}

\usepackage{microtype}
\usepackage{xcolor}
\usepackage{fancyhdr}

\usepackage{times}
\usepackage{latexsym}

\usepackage{CJKutf8}
\usepackage{booktabs}
\usepackage{graphicx}
\usepackage{subfigure}
\usepackage{comment}
\usepackage{xspace}
\usepackage{adjustbox}
\usepackage{mathtools}
\usepackage{bbm}
\usepackage{multirow}
\usepackage{amssymb}
\usepackage{amsmath}
\usepackage{verbatim}
\usepackage{amsfonts}
\usepackage{bm}
\usepackage[normalem]{ulem}
\usepackage{wrapfig}
\usepackage{bbding}

\title{Plug-and-Play Knowledge Injection for Pre-trained Language Models}

\author{Zhengyan Zhang$^{1*}$, Zhiyuan Zeng$^{1*}$, Yankai Lin$^{2,3}$, Huadong Wang$^{1}$, Deming Ye$^{1}$\\ 
\textbf{Chaojun Xiao$^{1}$, Xu Han$^{1\dagger}$, Zhiyuan Liu$^{1,4,5\dagger}$, 
Peng Li$^{6}$, Maosong Sun$^{1,4}$, Jie Zhou$^{7}$}\\
\textsuperscript{\rm 1}NLP Group, DCST, IAI, BNRIST, Tsinghua University, Beijing \\
\textsuperscript{\rm 2}Gaoling School of Artificial Intelligence, Renmin University of China, Beijing \\
\textsuperscript{\rm 3} Beijing Key Laboratory of Big Data Management and Analysis Methods \\     
\textsuperscript{\rm 4}International Innovation Center of Tsinghua University, Shanghai \textsuperscript{\rm 5} Quan Cheng Laboratory\\
\textsuperscript{\rm 6} Institute for AI Industry Research (AIR), Tsinghua University, China \\        
 \textsuperscript{\rm 7} Pattern Recognition Center, WeChat AI, Tencent Inc \\ 
\texttt{\{zy-z19,zengzy20\}@mails.tsinghua.edu.cn} \texttt{\{hanxu2022,liuzy\}@tsinghua.edu.cn}\\
}

\begin{document}
\maketitle
\begin{abstract}
Injecting external knowledge can improve the performance of pre-trained language models (PLMs) on various downstream NLP tasks. However, massive retraining is required to deploy new knowledge injection methods or knowledge bases for downstream tasks. In this work, we are the first to study how to improve the flexibility and efficiency of knowledge injection by reusing existing downstream models. To this end, we explore a new paradigm \textit{plug-and-play knowledge injection}, where knowledge bases are injected into frozen existing downstream models by a \textit{knowledge plugin}. Correspondingly, we propose a plug-and-play injection method \textit{map-tuning}, which trains a mapping of knowledge embeddings to enrich model inputs with mapped embeddings while keeping model parameters frozen. Experimental results on three knowledge-driven NLP tasks show that existing injection methods are not suitable for the new paradigm, while map-tuning effectively improves the performance of downstream models. Moreover, we show that a frozen downstream model can be well adapted to different domains with different mapping networks of domain knowledge. 
Our code and models are available at \url{https://github.com/THUNLP/Knowledge-Plugin}.
\end{abstract}

\section{Introduction}

{\let\thefootnote\relax\footnotetext{$^*$ Equal contribution}}
{\let\thefootnote\relax\footnotetext{$^\dagger$ Corresponding authors}}

Recent years have witnessed rapid development in enhancing pre-trained language models (PLMs) with various external knowledge bases, i.e., knowledge injection for PLMs~\cite{SenseBERT,LIMIT-BERT,ERNIE,KnowBERT,COMET,KnowledgeStory}.
Knowledge injection improves the performance of PLMs on a wide range of tasks such as information extraction~\cite{K-BERT,KEPLER}, question answering~\cite{WKLM,K-Adapter}, and text generation~\cite{KGPT}.

\begin{figure}
    \centering
    \includegraphics[width=\linewidth]{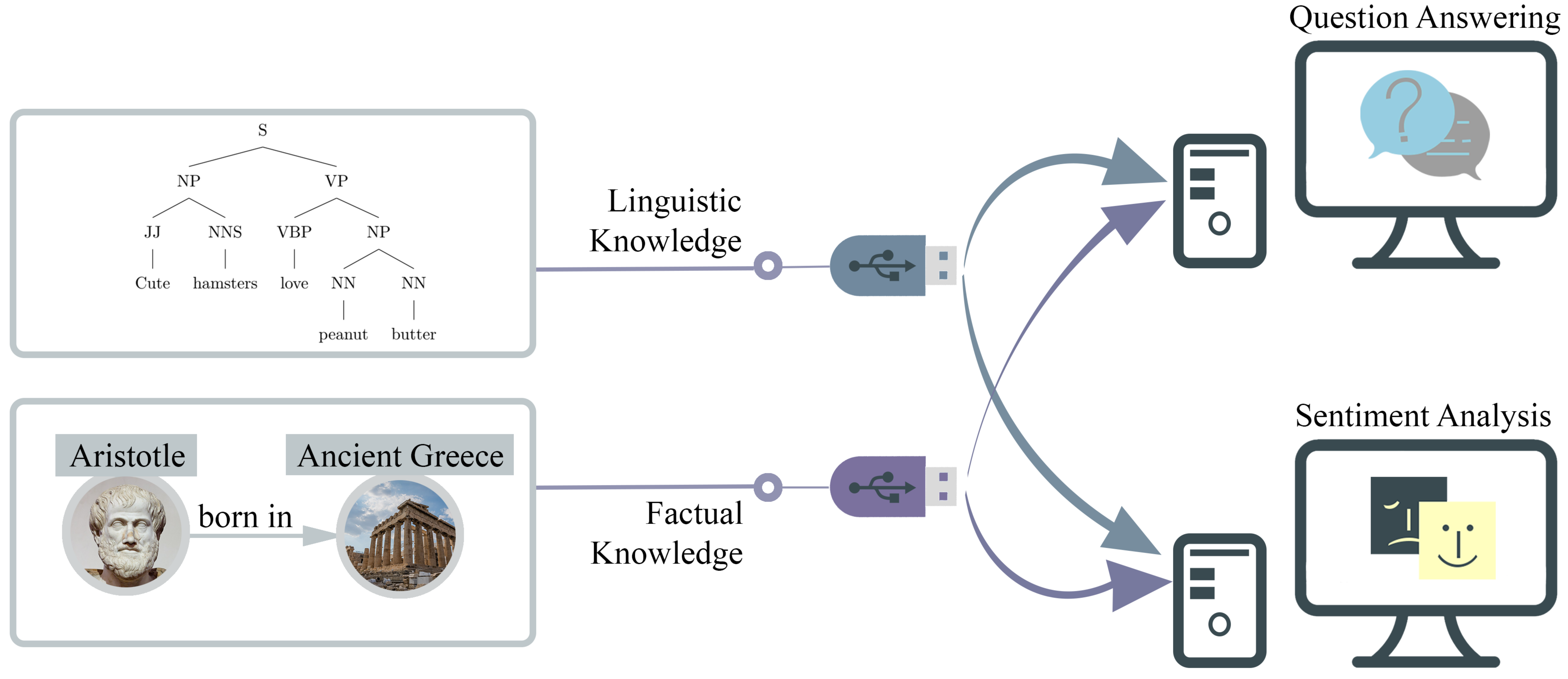}
    \caption{Illustration of plug-and-play knowledge injection, where knowledge bases and models are decoupled.}
    \label{fig:example}
    \vspace{-1.5em}
\end{figure}

Existing injection methods commonly inject knowledge by knowledge-aware pre-training or fine-tuning~\cite{KnowBERT,LUKE,K-BERT,K-Adapter}.
However, rarely studied is how to inject knowledge into a downstream model that is already adapted to a specific task.
If we want to apply a new knowledge injection method to enhance models on a specific task, we have to discard task-specific downstream models and retrain them.
In addition, one downstream model working with multiple knowledge bases requires retraining itself to inject each knowledge base.
Retraining models is time-consuming and resource-intensive, leading to the need for a \textbf{flexible and efficient} injection paradigm.

Toward flexible and efficient injection, we explore a novel paradigm \textit{plug-and-play knowledge injection}, where knowledge bases are injected into frozen existing downstream models by knowledge modules. The knowledge module bridges the knowledge base and the model, and we call it a plugin vividly.
Under this paradigm, a downstream model would have multiple plugins, each corresponding to a combination of an injection method and a knowledge base, which ensures flexibility.
Moreover, knowledge plugins should be small enough to ensure efficiency.
Intuitively, as shown in Figure~\ref{fig:example}, we treat models and knowledge bases as computers and flash disks, respectively.

In this work, we study two settings for the plug-and-play knowledge injection paradigm. 
The first is \textit{general plug-and-play knowledge injection}, aiming to inject knowledge into all downstream models (trained from a particular PLM) by a general plugin without any task-specific training.
In this setting, all downstream models share exactly one plugin for one combination of an injection method and a knowledge base.
The second is \textit{task-specific plug-and-play knowledge injection}, where knowledge plugins are trained to better adapt to downstream tasks while keeping downstream models frozen.

By our pilot study, we find that existing methods~\cite{E-BERT,PELT,K-Adapter,RAG} that can be used directly can not be well applied to the plug-and-play injection paradigm.
To this end, we propose \textit{map-tuning}, a preliminary exploration of learning knowledge plugins. 
Specifically, we train a lightweight mapping network that augments model inputs with mapped knowledge representations, e.g., TransE~\cite{TransE}.
To meet the general and task-specific injection requirements, we design general map-tuning and task-specific map-tuning, respectively. 
General map-tuning adopts language modeling as its objective to learn knowledge plugins and seeks better generalizability.
Task-specific map-tuning adopts task targets for plugin learning and seeks better task adaptation.

We use three typical knowledge-driven NLP tasks to evaluate our plug-and-play knowledge injection, including relation classification~\cite{FewRel-1}, entity typing~\cite{Wiki-ET}, and question answering~\cite{EntityQuestions}. The experimental results show that: 
(1) after adapting PLMs to downstream tasks through full-parameter fine-tuning or parameter-efficient tuning, also known as delta tuning~\cite{DBLP:journals/corr/abs-2107-13586,Delta}, injecting knowledge into these downstream models by general map-tuning leads to performance improvements in almost all cases; 
(2) using task-specific map-tuning to inject domain knowledge further enables a frozen downstream model to work well in different domains.
We hope our contribution can draw more attention to the plug-and-play knowledge injection paradigm and inspire more future research.

\section{Plug-and-Play Knowledge Injection}

\begin{figure*}
\centering
\includegraphics[width=0.9\linewidth]{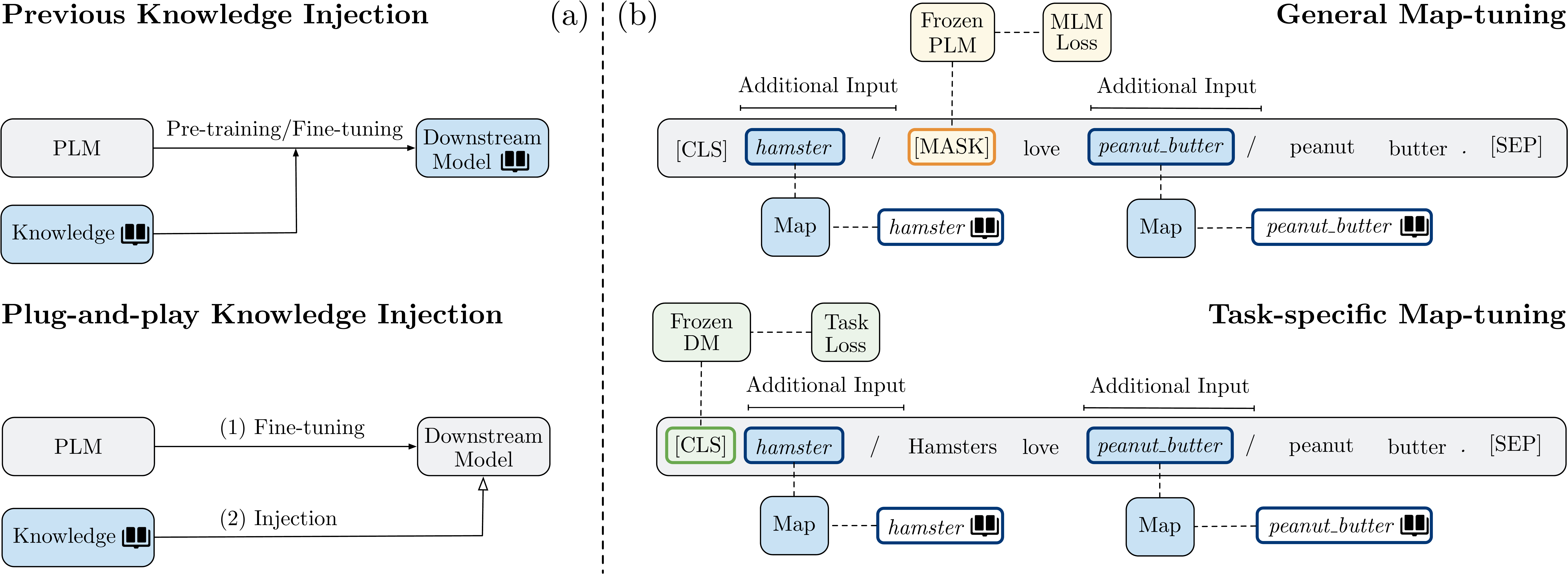}
\caption{Left: Comparisons between previous paradigms and our proposed plug-and-play paradigm. Right: Two ways for map-tuning. The input text is ``Hamsters love peanut butter.''. ``DM'' refers to ``downstream model''.}
\label{fig:model}
\vspace{-0.5em}
\end{figure*}

\paragraph{Paradigm Description.} Given a downstream model $\mathcal{D}$ trained on a downstream task with a PLM $\mathcal{P}$ as the backbone, we intend to improve its performance on this task by incorporating an extra knowledge base $\mathcal{B}$ and freezing $\mathcal{D}$'s parameters, for which we need to train a knowledge plugin $\mathcal{M}$. Note that neither pre-training nor fine-tuning trains the model $\mathcal{D}$ to cooperate with $\mathcal{B}$ or $\mathcal{M}$.

\paragraph{Two Injection Settings.}
As shown in Figure~\ref{fig:model}(a), plug-and-play knowledge injection decouples knowledge injection from model training, which is different from existing paradigms. For general plug-and-play knowledge injection, $\mathcal{M}$ is obtained based on only $\mathcal{P}$ and $\mathcal{B}$, and then it is directly plugged into all downstream models, $\mathcal{D}_1, \mathcal{D}_2, \ldots$, without any additional training. For task-specific plug-and-play knowledge injection, it is allowed to train $\mathcal{M}_1, \mathcal{M}_2, \ldots$ for $\mathcal{D}_1, \mathcal{D}_2,\ldots$ respectively while keeping $\mathcal{D}_1, \mathcal{D}_2,\ldots$ frozen.

\paragraph{Challenges.}
The general plug-and-play knowledge injection poses serious challenges to methods designed for it. 
$\mathcal{M}$ is expected to improve the performance of $\mathcal{D}$, yet $\mathcal{D}$ has never seen $\mathcal{M}$ or been seen by $\mathcal{M}$ during training. The only prior condition is that $\mathcal{P}$ and $\mathcal{B}$ are visible during training $\mathcal{M}$. Therefore, the designed methods for general injection need to endow $\mathcal{M}$ with enough generalizability such that $\mathcal{M}$ can adapt to unknown $\mathcal{D}_1, \mathcal{D}_2,\ldots$ well. Even though the knowledge base $\mathcal{B}$ may have rich knowledge, without a good adaptation of $\mathcal{M}$, useful information brought to D will be less than disruptive noise.

The task-specific plug-and-play knowledge injection relaxes the restrictions, where $\mathcal{M}_i$ is allowed to be trained with frozen $\mathcal{D}_i$.
Compared to injection during fine-tuning, the training of $\mathcal{M}_i$ should be fast and the parameter number of $\mathcal{M}_i$ should be small compared to that of $\mathcal{D}_i$. Otherwise, the methods would be meaningless.
Hence, it requires simple and efficient architecture designs and informative training objectives for $\mathcal{M}_i$.

\paragraph{Potentiality of Using Existing Methods.} 
Few existing knowledge injection methods can be directly used for general plug-and-play knowledge injection.
We summarize the existing knowledge injection methods\footnote{These methods are originally designed for injecting knowledge during pre-training or fine-tuning.} that have the possibility to be used for general plug-and-play knowledge injection as follows.
(1) Embedding-based methods: E-BERT~\cite{E-BERT} and PELT~\cite{PELT} build an entity embedding lookup table in the representation space of token embeddings and combine entity embeddings with token embeddings to construct input embeddings. 
(2) Retrieval-based methods: RAG~\cite{RAG} retrieves plain text from knowledge bases and augments the original input text with the plain text as injected knowledge.
(3) Adapter-based methods: K-Adapter~\cite{K-Adapter} computes knowledgeable representations based on the outputs of the downstream models accompanied by knowledgeable adapters, which are trained with frozen PLMs and plugged into all downstream models.
Even though these methods may bring knowledge without training PLMs, it is unclear whether they work well in the plug-and-play knowledge injection paradigm, i.e., whether the knowledge brought by them is utilizable for downstream models that have never learned how to use these methods.

\section{Map-Tuning}

In this section, we first present the overall framework of map-tuning, which is designed for plug-and-play knowledge injection. Then, we show how to use it for general injection and task-specific injection, where the methods are called general map-tuning and task-specific map-tuning respectively.

\subsection{Overall Framework}

We target knowledge bases consisting of a set of entities and structured or unstructured knowledge about these entities. To utilize such a knowledge base $\mathcal{B}$, we assume a knowledge representation model $\mathcal{K}$ to assign each entity $e$ an entity embedding $\mathbf{e} \in \mathbb{R}^{d_{\mathrm{KE}}}$, where $d_{\mathrm{KE}}$ is the dimension of entity embeddings. 
Map-tuning injects knowledge by mapping knowledge representations into the space of token embeddings and using the mapped representations as additional inputs, which is also adopted by~\citet{E-BERT,PELT}.

Specifically, given an input text, we first match the entity mentions in the text with the entities in $\mathcal{B}$. The input text is denoted by $\{w_1, w_2, \dots, w_n\}$, where $w_i$ is the $i$-th token and $n$ is the number of tokens in the input text. We use a triple $(e, l, r)$ to represent a mention span, where $e$ is the matched entity, $l$ and $r$ are the left and right token indices of the mention span. The corresponding mention span is $\{ w_l, w_{l + 1}, \dots, w_r \}$. Assume there are $m$ entities in the text, $(e_1, l_1, r_1), (e_2, l_2, r_2), \dots, (e_m, l_m, r_m)$, where $1 \leq l_1 \leq r_1 < l_2 \leq r_2 < \dots < l_m \leq r_m \leq n$. The original sequence of input embeddings are $\{\mathbf{w}_1, \mathbf{w}_2, \dots, \mathbf{w}_n\}$, where $\mathbf{w}_i \in \mathbb{R}^{d_{\mathrm{PLM}}}$ is the $i$-th token embedding and $d_{\mathrm{PLM}}$ is the dimension of token embeddings. Then, we map each entity embedding $\mathbf{e}_i$ to $\mathcal{M}(\mathbf{e}_i) \in \mathbb{R}^{d_{\mathrm{PLM}}}$ by a mapping network $\mathcal{M}$. Finally, we replace $\{ \mathbf{w}_{l_i}, \mathbf{w}_{l_i + 1}, \dots, \mathbf{w}_{r_i} \}$ with $\{\mathcal{M}(\mathbf{e}_i), \mathbf{/}, \mathbf{w}_{l_i}, \dots, \mathbf{w}_{r_i}\}$ for every $(e_i,l_i,r_i)$ to construct a new input sequence. Note that $\mathbf{/}$ is the token embedding of ``/''.

\subsection{General Map-tuning}

General map-tuning aims to train a mapping network $\mathcal{M}$ based on $\mathcal{P}$ and $\mathcal{K}$. 
It requires $\mathcal{M}$ to have enough generalizability to handle different downstream tasks because $\mathcal{M}$ will be plugged into all downstream models. Hence, we train $\mathcal{M}$ with a general pre-training task while plugging it into $\mathcal{P}$, such as language modeling, which has been shown to be an unsupervised multi-task learning~\cite{GPT-2}. 
We freeze the parameters of $\mathcal{P}$ and only train the mapping network $\mathcal{M}$ to meet the requirement of plug-and-play knowledge injection.

We adopt a variant of Masked Language Model (MLM)~\cite{BERT}, named Mention-Masked Language Modeling (MMLM), as the task for training $\mathcal{M}$. 
According to our observation in the preliminary experiments, the prediction of most tokens requires only language ability instead of external knowledge, such as that of some stop words, while the prediction of entity mentions relies on external knowledge more often.
Hence, as shown in Figure~\ref{fig:model}(b), we randomly mask only entity mentions\footnote{If the number of entity mentions is small, we can choose to cover all the masking combinations.} in the input text to ensure that the mapping network is trained sufficiently and the mapped embeddings are well utilized in the PLM. In this way, the ability of PLMs to predict masked entity mentions is enhanced by the mapped embeddings of both the masked entity and other entities in the context.
We mask all tokens of a masked entity mention, and the MMLM loss is the same as the original MLM loss~\cite{BERT}.

After general map-tuning, $\mathcal{M}$ can be used for the general plug-and-play injection. Although the mapping network $\mathcal{M}$ was not trained with any downstream model $\mathcal{D}$ before, we can directly plug $\mathcal{M}$ into each $\mathcal{D}$.

\subsection{Task-specific Map-tuning}

Task-specific map-tuning aims to adapt a mapping network $\mathcal{M}$ for a given downstream model $\mathcal{D}$. We freeze the parameters of $\mathcal{D}$ and train the mapping network $\mathcal{M}$ on the downstream task, whose procedure is shown in Figure~\ref{fig:model}(b). The training objective is identical to the original objective of this task. If the knowledge representations provide useful information for this task, the mapping network will learn to extract this information and to make it recognizable to the downstream model $\mathcal{D}$. Note that the mapping network can not only be trained from scratch, but can also be initialized with a mapping network learned with general map-tuning, which could provide a good starting point.

\section{Experiments}

\subsection{Experimental Setups}

\textbf{Training Methods of Downstream Models.} We adopt BERT$_\textrm{base}$~\cite{BERT} as the backbone PLM in the experiments and consider four training methods for its adaptation to downstream tasks. Besides vanilla full-model fine-tuning, we also consider three parameter-efficient tuning (PET) methods, which have been becoming increasingly important in the era of large-scale PLMs~\cite{DBLP:journals/corr/abs-2107-13586}. 
As resource-saving are both plug-and-play knowledge injection and PET, it is meaningful to apply this paradigm to downstream models trained by PET methods in the resource-limited scenario.
(1) \textbf{Fine-tuning} optimizes all the parameters of a PLM with the task objective following the original BERT. (2) \textbf{LoRA}~\cite{LoRA} freezes most PLM parameters and represents the weight update during model training with a low-rank decomposition.
(3) \textbf{Adapter}~\cite{Adapter} injects additional adapter networks with the PLM parameters frozen. (4) \textbf{BitFit}~\cite{BitFit} only optimizes the parameters of bias vectors and freezes the rest parameters. 
The hyper-parameters are reported in Appendix~\ref{sec:hyper-parameters}.

\textbf{Downstream Tasks.} We evaluate methods under the plug-and-play knowledge injection paradigm on three kinds of knowledge-driven NLP tasks including relation classification, entity typing, and question answering. 
For relation classification, which requires models to classify the relation between two entities given a context, we experiment on both few-shot and full-data settings. In the few-shot setting, we aim to evaluate model performance on long-tail relations whose training instances are not sufficient. Specifically, we use FewRel 1.0~\cite{FewRel-1} and FewRel 2.0~\cite{FewRel-2}.\footnote{We randomly sample 5000 instances from test data of FewRel 1.0 and FewRel 2.0 respectively for fast evaluation. Note that the test data is not publicly released and we get the data from the authors. We experiment with full test data of FewRel 1.0 on the official leaderboard in Section~\ref{sec:efficiency}.} In the full-data setting, we evaluate models on Wiki80~\cite{opennre}, which contains 80 relation types from Wikidata, and follow the data split of~\citet{ERNIE}. 
For entity typing, which requires models to classify the type of an entity given a context, we evaluate models on Wiki-ET~\cite{Wiki-ET} containing 68 entity types from Freebase. 
For question answering, we evaluate models on EntityQuestions~\cite{EntityQuestions}, an open-domain QA dataset consisting of entity-centric questions.
We use knowledge-enhanced models to directly answer questions without retrieving related documents.
We report accuracy on relation classification and question answering, and F1 score on entity typing.

\textbf{Knowledge Bases.} 
We use Wikidata and UMLS\footnote{UMLS represents the Unified Medical Language System, which is the trademark of U.S. National Library of Medicine.} as our external knowledge bases for the Wikipedia domain and PubMed\footnote{\url{https://pubmed.ncbi.nlm.nih.gov/}} domain, respectively. Specifically, we use the Wikidata triples provided by~\citet{ERNIE} and the Wikidata entity descriptions provided by~\citet{KEPLER}.
To avoid information leakage in the relation classification task, we remove the triples appearing in the datasets from these knowledge bases.
We adopt TransE~\cite{TransE} as our knowledge representation model and the dimension of knowledge embeddings is set to $100$.

\textbf{Evaluated Existing Methods.} We evaluate existing methods that can be applied to general plug-and-play knowledge injection. (1) \textbf{E-BERT}~\cite{E-BERT} also obtains a mapping network to transform knowledge embeddings. 
Different from map-tuning, E-BERT builds the connection between the vocabulary and entities by string matching, and then make the mapped knowledge embeddings close to their corresponding token embeddings. 
In this work, E-BERT uses the same TransE embeddings as map-tuning instead of wikipedia2vec for fair comparisons.
(2) \textbf{PELT}~\cite{PELT} aggregates the output representations of a specific entity in multiple contexts to build the entity representation. Then, the entity representation can be appended to the model input without any mapping because the input space and output space are the same for most PLMs. 
The entity-related context can be treated as an external textual knowledge base.
(3) \textbf{Retrieval Augmentation (RA)} is to augment input texts with additional retrieved unstructured knowledge, such as RAG~\cite{RAG} and REALM~\cite{REALM}. In this work, we retrieve the entity descriptions from Wikidata5M and append them to the input texts.
(4) \textbf{K-Adapter}~\cite{K-Adapter} implicitly stores knowledge in the parameters of adapter networks. We follow the original procedure of K-Adapter while keeping the parameters of PLMs and adapters frozen.\footnote{This procedure still requires the training of the final fully connected layer, which does not strictly meet the setting of general plug-and-play Injection.}

\textbf{Details of Map-tuning.} The architecture of the mapping network is simply an affine transformation $\mathbf{W}\mathbf{e}+\mathbf{b}$, where $\mathbf{W} \in \mathbb{R}^{d_\textrm{PLM}\times d_\textrm{KE}}$ and $\mathbf{b} \in \mathbb{R}^{d_\textrm{PLM}}$. In this work, the parameter amount of the mapping network is $768\times 128+768<0.1\text{M}$. For Mention-Masked Language Modeling, we use the raw texts of Wiki20M~\cite{wiki20m}, which is sampled from the Wikipedia corpus and provides the annotations of entity linking.
The total size is around 300MB, much smaller than common pre-training corpora. Since map-tuning only aims to adapt the mapping network for a PLM, it does not require much training data. 
We train the mapping network for $5$ epochs, which costs only $12$ hours on an NVIDIA Tesla V100. 
General map-tuning essentially builds an entity embedding lookup table. To evaluate its quality, we evaluate it in the traditional injection during fine-tuning paradigm as a preliminary experiment. To be more specific, we fine-tune the PLMs on downstream tasks, during which the mapping network is plugged into them. The details are in Appendix~\ref{sec:fine-tuning}. We find that map-tuning consistently outperforms E-BERT and PELT in the traditional paradigm, which also builds entity embedding lookup tables.

\subsection{General Plug-and-Play Injection}

In this subsection, we evaluate knowledge injection methods in the setting of general plug-and-play knowledge injection, where we directly plug knowledge modules into downstream models without any training.
The results are reported in Table~\ref{tab:general-injection-results}.

\begin{table*}[h]
\centering
\scriptsize
\begin{tabular}{ll|llll|l|l|l}
\toprule
\multirow{2}{*}{Method} & \multirow{2}{*}{Injection} & \multicolumn{4}{c|}{FewRel 1.0} & \multirow{2}{*}{Wiki80} & \multirow{2}{*}{Wiki-ET}  & \multirow{2}{*}{EntityQuestions} \\ 
&            &     \multicolumn{1}{c}{5-1}     &      \multicolumn{1}{c}{5-5}     &     \multicolumn{1}{c}{10-1}     &     \multicolumn{1}{c|}{10-5}     &         &  \\ 
\midrule    \multirow{6}{*}{Fine-tuning} & $-$          &    91.0    &    95.1    &    85.4    &    90.8    &    86.1    & 77.5  & 41.7 \\
& E-BERT     & 91.0 ($+$0.0) & 95.0 ($-$0.1) & 86.5 ($+$1.1) & 90.5 ($-$0.3) & 85.4 ($-$0.7) & 77.0 ($-$0.5)  &  42.9 ($+$1.2) \\
& PELT       & 90.5 ($-$0.5) & 94.8 ($-$0.3) & 85.3 ($-$0.1) & 89.8 ($-$1.0) & 85.0 ($-$1.1) & 76.8 ($-$0.7)   &  46.8 ($+$5.1) \\
& RA        & 91.5 ($+$0.5) & 95.5 ($+$0.4) & 85.8 ($+$0.4) & \textbf{91.7} ($+$\textbf{0.9}) & 85.9 ($-$0.2)  & 76.7 ($-$0.8)   &  \textbf{69.5} ($+$\textbf{27.8}) \\
& K-Adapter  & 88.6 ($-$2.4) & 94.5 ($-$0.6) & 82.3 ($-$3.1) & 89.9 ($-$0.9) & 86.0 ($-$0.1) & \textbf{77.8} ($+$\textbf{0.3})    &   39.2 ($-$2.5) \\
& Map-tuning & \textbf{92.6} ($+$\textbf{1.6}) & \textbf{95.6} ($+$\textbf{0.5}) & \textbf{88.1} ($+$\textbf{2.7}) & 91.2 ($+$0.4) & \textbf{86.7} ($+$\textbf{0.6}) & 76.6 ($-$0.9)   & 49.0 ($+$7.3)  \\ \midrule
\multirow{6}{*}{LoRA}        & $-$          &    90.7    &    95.1    &    84.9    &    91.2    &    85.3    & 77.5    &   42.4    \\
& E-BERT     & 90.7 ($+$0.0) & 95.2 ($+$0.1) & 85.4 ($+$0.5) & 90.4 ($-$0.8) & 83.7 ($-$1.6) & 77.6 ($+$0.1)   &  44.0 ($+$1.6)   \\
& PELT       & 89.9 ($-$0.8) & 94.8 ($-$0.3) & 84.6 ($-$0.3) & 89.8 ($-$1.4) & 83.1 ($-$2.2) & 77.5 ($+$0.0)    & 47.7 ($+$5.3)  \\
& RA        & 91.3 ($+$0.6) & 95.8 ($+$0.7) & 85.0 ($+$0.1) & \textbf{92.5} ($+$\textbf{1.3}) & 83.8 ($-$1.5) & 76.8 ($-$0.7)   &   47.7 ($+$5.3) \\
& K-Adapter  & 90.0 ($-$0.7) & 94.8 ($-$0.3) & 83.4 ($-$1.5) & 89.1 ($-$2.1) & 85.0 ($-$0.3) & 77.3 ($-$0.2) & 41.1 ($-$1.3)\\
& Map-tuning & \textbf{92.3} ($+$\textbf{1.6}) & \textbf{96.0} ($+$\textbf{0.9}) & \textbf{87.4} ($+$\textbf{2.5}) & 91.9 ($+$0.7) & \textbf{85.8} ($+$\textbf{0.5}) & \textbf{78.3} ($+$\textbf{0.8})  &  \textbf{49.6} ($+$\textbf{7.2}) \\ \midrule
\multirow{6}{*}{Adapter}     & $-$          &    91.2    &    95.2    &    86.2    &    91.1    &    85.7    & 77.5   &  43.6  \\
& E-BERT     & 91.3 ($+$0.1) & 95.4 ($+$0.2) & 86.9 ($+$0.7) & 91.6 ($+$0.5) & 84.4 ($-$1.3) & 78.4 ($+$0.9)  & 45.1 ($+$1.5) \\
& PELT       & 91.0 ($-$0.2) & 95.4 ($+$0.2) & 86.3 ($+$0.1) & 91.3 ($+$0.2) & 84.3 ($-$1.4) & 77.9 ($+$0.4)   & 48.4 ($+$4.8) \\
& RA        & 91.7 ($+$0.5) & 95.5 ($+$0.3) & 85.8 ($-$0.4) & \textbf{92.3} ($+$\textbf{1.2})  & 85.0 ($-$0.7)  & 76.8 ($-$0.7)   &  42.9 ($-$0.7)  \\
& K-Adapter  & 89.9 ($-$1.3) & 94.7 ($-$0.5) & 83.6 ($-$2.6) & 90.0 ($-$1.1) & \textbf{85.9} ($+$\textbf{0.2}) &  77.7 ($+$0.2)  & 41.5 ($-$2.1)\\
& Map-tuning & \textbf{92.6} ($+$\textbf{1.4}) & \textbf{95.8} ($+$\textbf{0.6}) & \textbf{88.2} ($+$\textbf{2.0}) & 91.8 ($+$0.7) & \textbf{85.9} ($+$\textbf{0.2}) & \textbf{79.2} ($+$\textbf{1.7})    &  \textbf{50.8} ($+$\textbf{7.2})\\ \midrule
\multirow{6}{*}{BitFit}      & $-$          &    89.2    &    94.8    &    83.0    &    90.0    &    82.7    & 77.1   & 41.3\\
& E-BERT     & 88.7 ($-$0.5) & 94.5 ($-$0.3) & 83.5 ($+$0.5) & 89.6 ($-$0.4) & 81.3 ($-$1.4) & 77.2 ($+$0.1)   &  42.3 ($+$1.0) \\
& PELT       & 88.2 ($-$1.0) & 94.3 ($-$0.5) & 80.9 ($-$2.1) & 88.3 ($-$1.7) & 80.3 ($-$2.4) & 77.6 ($+$0.5)    &  46.7 ($+$5.4) \\
& RA        & 89.5 ($+$0.3) & 95.2 ($+$0.4) & 82.7 ($-$0.3) & \textbf{91.1} ($+$\textbf{1.1}) & 81.8 ($-$0.9) & 74.0 ($-$3.1)    & 33.9 ($-$7.4)  \\
& K-Adapter  & 86.4 ($-$2.8) & 93.7 ($-$1.1) & 78.8 ($-$4.2) & 87.5 ($-$2.5) & 81.5 ($-$1.2) &  77.2 ($+$0.1)   &  40.7 ($-$0.6)\\
& Map-tuning & \textbf{90.4} ($+$\textbf{1.2})  & \textbf{95.5} ($+$\textbf{0.7}) & \textbf{85.2} ($+$\textbf{2.2}) & 90.8 ($+$0.8) & \textbf{83.7} ($+$\textbf{1.0}) & \textbf{78.0} ($+$\textbf{0.9})  &  \textbf{48.4} ($+$\textbf{7.1}) \\ \bottomrule
\end{tabular}
\caption{Results of general plug-and-play injection. We adapt BERT$_\textrm{base}$ to these datasets with four different training methods. There are five different injection methods. E-BERT, PELT, and Map-tuning utilize entity representations. RA utilizes entity descriptions as additional text input. K-Adapter utilizes the knowledge implicitly stored in the adapter network. Note that downstream models and injection models are trained separately. N-K indicates the N-way K-shot configuration. We boldface the best result for each training method.}
\vspace{-1em}
\label{tab:general-injection-results}
\end{table*}

From this table, we have four observations:
(1) All of the four existing methods can not consistently improve the performance of downstream models. In most cases, injecting these knowledge modules degrades the model performance, often to a large degree. It empirically proves that \textbf{the setting of general plug-and-play injection is challenging} and these four methods are not suitable in this setting. The knowledge provided by these methods can not be directly used, so they are basically disruptive noise to the downstream models. 
(2) Our proposed \textbf{general map-tuning achieves consistent improvement} on almost all downstream models, suggesting that the mapping network effectively transforms knowledge embeddings into the space of token embeddings and the mapped embeddings can be directly used by downstream models. 
We highlight the importance of Mention-Masked Language Modeling, which provides sufficient training instances for general map-tuning, while the matched entity-token pairs for E-BERT are insufficient for training the mapping network.
(3) Intuitively, general map-tuning may work better with PET methods than with full-model fine-tuning because PET methods change much fewer parameters from the PLM and general map-tuning is trained based on the PLM. 
In fact, the performance improvement brought to models trained by full-model fine-tuning is comparable to that of PET methods. It demonstrates that map-tuning is a promising method regardless of the training methods of downstream models.
(4) Remarkably high is the performance improvement brought by RA to fine-tuned BERT on EntityQuestions. We observe that the retrieved entity description contains the exact answer as a substring for 62.19\% of instances in the test set, and we remove these instances and report the result in Table~\ref{tab:remove_stringmatch}. We find that RA still gets a slightly higher performance than map-tuning does for fine-tuned BERT, but brings a significant performance drop to other downstream models, while map-tuning brings consistent performance improvement to all downstream models.
It suggests that fine-tuned BERT has the surprising generalization ability to extract a substring in the additional context as the answer, and even to reveal the answer hidden in the additional context without string matches.
On the contrary, other downstream models are not able to reveal the hidden answer.
Thus, it is worth investigating RA with pluggable knowledge modules to stably provide information for different downstream models, rather than directly appending unstructured text to model inputs.

\subsection{Task-specific Plug-and-Play Injection}
\label{sec:task-specific}

Since map-tuning achieves the best performance in the general plug-and-play injection setting, we further evaluate it in the setting of task-specific plug-and-play injection, where we train mapping networks based on downstream models with task objectives. If we have already conducted general map-tuning on a PLM, we can initialize the network with the general mapping network. Otherwise, we have to train the network from scratch.

\begin{table}[t]
\centering
\scriptsize
\begin{tabular}{l|c|c|c}
\toprule
Method & Wiki80 & Wiki-ET & EntityQuestions \\\midrule
Fine-tuning                                   & 86.1 & 77.5   &   41.7   \\
+ General Map-tuning                                  & 86.7 & 76.6   &   49.0   \\
\midrule
\multicolumn{3}{l}{+ Task-specific Map-tuning}  \\
\midrule
Train from Scratch             & 87.2 & 78.8   &   57.7   \\
Train from the General Map     & \textbf{87.8} & \textbf{78.9}    &  \textbf{58.9}   \\ \bottomrule  
\end{tabular}
\caption{Results of task-specific map-tuning. We train the mapping network from scratch or initialize the mapping network with the general mapping network.}
\label{tab:task-specific-map-tuning}
\vspace{-2em}
\end{table}

We first evaluate task-specific map-tuning on Wiki80 and Wiki-ET. The results are reported in Table~\ref{tab:task-specific-map-tuning}. From the table, we have two observations: (1) Task-specific map-tuning achieves better performance on these two datasets than general map-tuning does. It indicates that the mapping network extracts more informative knowledge for the specific task by task-specific training than the general one does. (2) If the general mapping network is available, it is recommended to use it to initialize the mapping network, which further improves the model performance.

\begin{table*}[]
\small
\centering
\begin{tabular}{lc|rrrr|rrrr}
\toprule
\multirow{2}{*}{Training Data} & \multirow{2}{*}{Map-tuning} & \multicolumn{4}{c|}{Source Domain} & \multicolumn{4}{c}{Target Domain} \\
&     & 5-1 & 5-5 & 10-1 & 10-5 & 5-1 & 5-5 & 10-1 & 10-5 \\
\midrule
Target Domain       & $-$  & 65.4 & 80.8 & 56.9 & 73.8 & 78.6 & 88.6 & 71.4 & 79.7 \\ \midrule
Multiple Domains & $-$  & 90.3 & 94.6 & 84.9 & 90.4 & \textbf{84.8} & \textbf{92.0} & \textbf{79.0} & \textbf{86.8} \\ 
\midrule
\multirow{2}{*}{Source Domain}        & $-$  & 91.0 & 95.1 & 85.4 & 90.8 & 76.7 & 88.2 & 69.1 & 81.5 \\
& \checkmark & \textbf{92.9} & \textbf{95.6} & \textbf{88.2} & \textbf{91.1} & 81.2 & 89.8 & 72.6 & 83.3 \\
\bottomrule
\end{tabular}
\caption{Results of domain adaptation. The source domain is Wikipedia from FewRel 1.0. The target domain is PubMed from FewRel 2.0. We compare task-specific map-tuning with fine-tuning on the target domain and multiple domains consisting of both source and target domains.}
\label{tab:domain-adaptation-results}
\vspace{-0.5em}
\end{table*}

Then, we evaluate task-specific map-tuning in domain adaptation, which is a more challenging setting.
In this setting, we aim to plug multiple knowledge bases into a single downstream model.
Specifically, a downstream model is trained on a source domain, and then we plug the knowledge modules of the target domain into it for domain adaptation.
Here, we use the relation classification datasets on the Wikipedia domain (FewRel 1.0) and the PubMed domain (FewRel 2.0). FewRel 1.0 is the source domain. FewRel 2.0 is the target domain. 
The knowledge base for FewRel 2.0 is UMLS.
Since the original FewRel 2.0 does not provide training instances, we rearrange FewRel 2.0 and have the following data split. As FewRel 2.0 has 25 relations, we separate 15 relations for training and development and the rest 10 relations are used for testing. 

From Table~\ref{tab:domain-adaptation-results}, we have two observations: (1) For the domain adaptation from Wikipedia to PubMed, map-tuning significantly improves the model performance (e.g., from 76.7 to 81.2 in 5-1) and achieves better performance than the model fine-tuned on PubMed domain (e.g., from 78.6 to 81.2 in 5-1). It suggests that it is promising to use map-tuning to introduce external knowledge for domain adaptation. (2) Multi-domain training degrades the model performance on the Wikipedia domain and maintains its performance on the PubMed domain while map-tuning does not degrade the performance on each domain. It indicates that the pluggable mapping networks are suitable for continual domain adaptation.

\subsection{Computational Efficiency}
\label{sec:efficiency}

\begin{table*}[]
\centering
\small
\begin{tabular}{lllll}
\toprule
 Input                 &   True label                & Injection & Predicted label & Logits \\
\midrule
\multirow{4}{*}{ \begin{tabular}[c]{@{}l@{}}
    \underline{Ernest Russell Lyon} was a\\ \uwave{flying officer} in 234 Squadron \\ of the Royal Air Force \\ during the Second World War.
    \end{tabular} } & \multirow{4}{*}{military\_rank} & \multirow{2}{*}{-} & \multirow{2}{*}{ \begin{tabular}[c]{@{}l@{}}
        occupation, military\_rank, \\ field\_of\_work
        \end{tabular} } & \multirow{2}{*}{8.0, 4.7, 3.3} \\
    &                   &  &  &  \\ \cline{3-5}
    &                   & General & \multirow{2}{*}{ \begin{tabular}[c]{@{}l@{}}
        military\_rank, field\_of\_work, \\ occupation
        \end{tabular} } & \multirow{2}{*}{6.3, 6.2, 3.9} \\
    &                   & Map-tuning &  &                \\
\midrule
\multirow{6}{*}{ \begin{tabular}[c]{@{}l@{}}
    He later enslaved Thor, then \\ captured the \underline{Wasp} and the other \\ \uwave{Avengers}.
    \end{tabular} } & \multirow{6}{*}{member\_of} & \multirow{2}{*}{-} & \multirow{2}{*}{ \begin{tabular}[c]{@{}l@{}}
        member\_of, parts, \\ characters 
        \end{tabular} } & \multirow{2}{*}{8.8, 5.0, 4.4} \\
& &  &  \\\cline{3-5}
& & General & \multirow{2}{*}{ \begin{tabular}[c]{@{}l@{}}
    characters, member\_of, \\  parts
    \end{tabular} } & \multirow{2}{*}{6.9, 6.6, 4.7} \\
 & & Map-tuning\\ \cline{3-5}
 & & Task-specifc & \multirow{2}{*}{ \begin{tabular}[c]{@{}l@{}}
    member\_of, parts, \\ characters 
    \end{tabular} } & \multirow{2}{*}{8.4, 5.7, 4.6}\\
 & & Map-tuning \\
\bottomrule
\end{tabular}
\caption{A case study for map-tuning on Wiki80. \underline{Underlines} and \uwave{wave lines} highlight head entities and tail entities respectively. We report the top 3 ranked predictions of different methods.}
\label{tab:case}
\vspace{-0.5em}
\end{table*}

We compare our proposed plug-and-play knowledge injection paradigm with previous knowledge injection paradigms on the time cost.
We evaluate the training time on an NVIDIA Tesla V100 and compare the model performance on the 10-way 1-shot setting of FewRel 1.0. ERNIE~\cite{ERNIE}, KEPLER~\cite{KEPLER}, and LUKE~\cite{LUKE} inject knowledge during pre-training. PELT~\cite{PELT} injects knowledge during fine-tuning. The results of ERNIE, KEPLER, LUKE, and PELT are taken from~\citet{PELT}. Map-tuning injects knowledge after fine-tuning.

\begin{figure}[t]
\centering
\includegraphics[width=\linewidth]{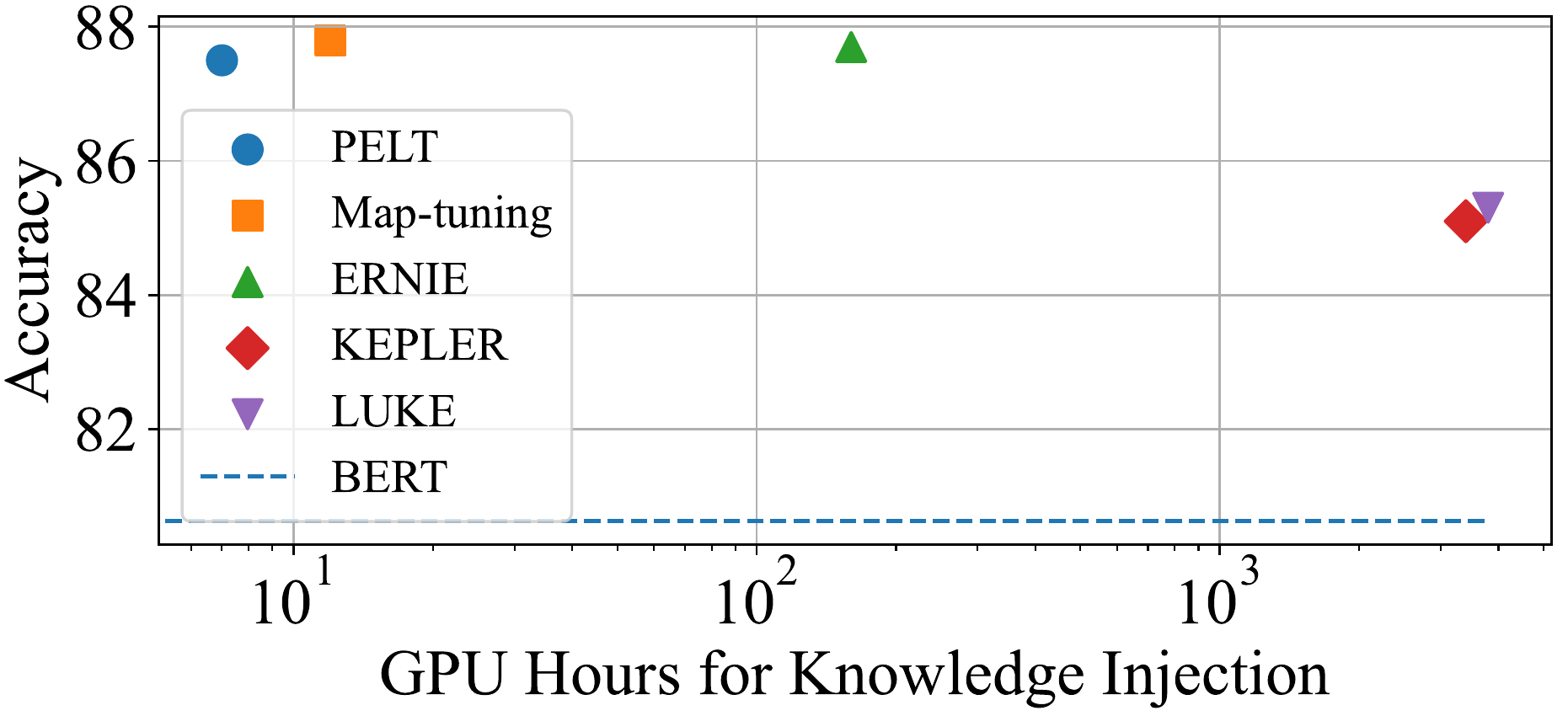}
\caption{Time cost of different knowledge injection methods on an NVIDIA Tesla V100 GPU.}
\label{fig:time-cost}
\vspace{-1em}
\end{figure}

The results are shown in Figure~\ref{fig:time-cost}. From this figure, we observe that the training time of map-tuning is much shorter than those methods under the paradigm of injecting during pre-training, and it is comparable to PELT. Besides, the performance of map-tuning is also competitive compared to previous knowledge injection methods.
Moreover, map-tuning only optimizes additional 0.1\% of parameters and we report the number of parameters optimized for different knowledge injection methods in Appendix~\ref{sec:parameters}. Plug-and-play knowledge injection has great potential to be comparable to previous paradigms w.r.t. task performance, while maintaining its innate flexibility and efficiency.

\subsection{Case Study}

We present a qualitative analysis of map-tuning in Table~\ref{tab:case}. In the first case, the original downstream model does not understand that ``flying officer'' is a military rank and wrongly predicts the relation as ``occupation''. With the general mapping network, which enriches the meaning of ``flying officer'', the model correctly predicts the relation.

The general mapping network, however, may be misleading in some cases. In the second case, it is easy for the original downstream model to recognize ``Wasp'' as a member of ``Avengers'' without any external knowledge since this fact could be inferred by the word ``other''. Compared to the external knowledge provided by the task-specific mapping network, coarse-grained is that provided by the general mapping network, because there is no additional training before the inference. As a result, the model wrongly recognizes ``Avengers'' as comic books instead of the fictional superhero team, and thus changes the correct model prediction. Task-specific map-tuning, which is further adapted to the task, corrects the prediction.

\section{Related Work}

To enhance PLMs with external knowledge, there are two mainstream paradigms: injection during pre-training and injection during fine-tuning~\cite{ke-survey}. 
For injection during pre-training, researchers usually construct new knowledge-aware objectives, such as entity prediction~\cite{DBLP:conf/emnlp/000200WW0LZ21}, entity discrimination~\cite{WKLM}, entity and relation discrimination~\cite{qin-etal-2021-erica}, and link prediction~\cite{KEPLER}. 
In this way, knowledge will be implicitly stored in the parameters of PLMs. Injection knowledge during pre-training can simultaneously improve performance on a range of downstream knowledge-driven tasks. However, the training cost of this paradigm is expensive. Taking the typical knowledge-enhanced PLMs LUKE~\cite{LUKE} and KEPLER~\cite{KEPLER} as an example, it takes more than 3,000 GPU hours to train them.

Injection knowledge during fine-tuning is a relatively lightweight paradigm, where external knowledge is often used to augment model inputs for specific tasks~\cite{GEAR,KagNet,KGAT,UnitedQA,KALA}. 
When injecting unstructured textual knowledge, some methods retrieve task-related information from external corpora to augment the original input text~\cite{DBLP:conf/emnlp/KarpukhinOMLWEC20,K-BERT}.
When using structured knowledge, such as knowledge graphs, existing methods usually apply knowledge representation learning methods~\cite{TransE,TransR} to encode structured knowledge into embeddings, and then fuse these knowledge embeddings with input token embeddings using knowledge injection methods~\cite{CoLAKE,CokeBERT,QA-GNN}.

In general, existing knowledge injection methods mainly target PLMs and adopt paradigms where knowledge and models are highly coupled.
Toward flexible and efficient injection, we study a new paradigm, plug-and-play knowledge injection, where we decouple models and knowledge sources, and then inject knowledge into downstream models without retraining the models.
This work is also related to parameter-efficient tuning~\cite{DBLP:journals/corr/abs-2107-13586,Delta} and plugins for large language models~\cite{xiao2023plug,PlugPLMControlGenration,DBLP:conf/emnlp/LauscherLG21,DBLP:conf/naacl/ChronopoulouPD22,yu2023augmentation,DBLP:journals/corr/abs-2305-08848,DBLP:conf/nips/AlayracDLMBHLMM22} while we are the first to study knowledge injection in a parameter-efficient and plug-and-play way.

\section{Conclusion}

In this work, we propose a new paradigm of injection toward flexible and efficient knowledge injection.
In this paradigm, downstream models can be enhanced with little computational cost, which benefits large amounts of models.
We first systematically evaluate existing knowledge injection methods and find that they are not suitable for plug-and-play injection. 
Then, we propose map-tuning for this paradigm, which effectively injects knowledge into downstream models to enhance them.

There are four promising directions for future investigation into plug-and-play knowledge injection. (1) How can we reduce the performance gap between methods for this novel paradigm and those for the previous injection paradigms, while maintaining superior flexibility and efficiency? (2) Besides factual knowledge, how can we effectively plug diverse knowledge bases, such as text corpora, voice, images, and even other PLMs? (3) After injecting the knowledge in a plug-and-play way, how can the PLMs do various types of complex reasoning based on the injected knowledge~\cite{LearnEntities}? (4) Can the plug-and-play knowledge injection methods for these sources be unified, so we can plug a combination of multiple sources? We hope this work can attract attention to and inspire research on these problems.

\section*{Limitations}
In this paper, we present a novel knowledge injection paradigm \emph{plug-and-play knowledge injection} for PLMs. We show existing methods can not be well applied to the new paradigm and propose \emph{map-tuning} as a preliminary exploration of methods.

The paradigm \emph{plug-and-play knowledge injection} has a limitation in terms of its assumption. It assumes that a PLM should be fine-tuned for downstream tasks. However, very large-scale PLMs can perform zero-shot learning or in-context learning on downstream tasks without being fine-tuned. Future work may extend the definition of the proposed paradigm to make it meaningful in these scenes.

The method \emph{map-tuning} has three limitations in terms of its applicability. Firstly, we did not evaluate map-tuning for PLMs pre-trained by other language modeling objectives (e.g., casual language modeling) besides MLM. As its spirit can be easily generalized to various language modeling objectives, we leave this evaluation as future work. Secondly, we did not evaluate whether the PLM can do complex reasoning (e.g., multi-hop reasoning) based on the knowledge injected by map-tuning. Thirdly, map-tuning is designed to plug structural fact knowledge. It is also meaningful to plug other diverse knowledge bases, including text corpora, voice, images, and even other PLMs, which are not covered by our work.

\section*{Acknowledgments}

This work is supported by the  National Key R\&D Program of China (No.2022ZD0116312), National Natural Science Foundation of China (No. 62236004).

\paragraph{Author Contributions} Zhengyan Zhang, Zhiyuan Zeng, Huadong Wang, and Deming Ye wrote the code and conducted the experiments. Zhengyan Zhang constructed the basic experimental framework including codes and datasets. Zhiyuan Zeng was in charge of plug-and-play and fine-tuning experiments. Huadong Wang and Deming Ye provided TransE and PELT embeddings respectively. Zhengyan Zhang and Zhiyuan Zeng contributed to the analysis experiments.
Zhengyan Zhang and Zhiyuan Zeng wrote the initial draft. Yankai Lin, Huadong Wang, Chaojun Xiao, Xu Han, and Zhiyuan Liu significantly edited and improved the paper.
Peng Li, Maosong Sun, and Jie Zhou provided valuable advice to the research.

\bibliography{anthology,custom}
\bibliographystyle{acl_natbib}

\newpage
\appendix

\section{Hyper-parameters}
\label{sec:hyper-parameters}

\subsection{Fine-tuning downstream PLMs}

We experiment with four training methods for the adaptation of PLMs on downstream tasks, which are Full-model fine-tuning, LoRA, Adapter, and BitFit. The embedding layer is frozen during training. We train all the models using AdamW with 10\% warming-up steps. We list our hyper-parameters in Table~\ref{tab:hyper-parameters-fine-tuning}.

\begin{table}[]
\centering
\tiny
\scalebox{0.8}{
\begin{tabular}{l|l|r|r|r|r}
\toprule
Method & Hyper-parameters & FewRel & Wiki80 & Wiki-ET  &  EntityQuestions   \\
\midrule
- & Sequence Length & 512 & 128 & 64   &   64 \\
\midrule
\multirow{3}{*}{Fine-tuning}
& Learning Rate & 2E-5 & 5E-5 & 1E-5   &   1E-4 \\
& Batch Size & 4 & 32 & 64   & 64 \\
& Training Step/Epoch & 3000 & 15 & 2  &   5     \\
\midrule
\multirow{4}{*}{LoRA}
& Learning Rate & 8E-4 & 2E-3 & 1E-3  &   5E-4 \\
& Batch Size & 4 & 64 & 64    &   64\\
& Training Step/Epoch & 3000 & 60 & 2   & 5    \\
& Rank & 32 & 32 & 4  & 4  \\
\midrule
\multirow{4}{*}{Adapter}
& Learning Rate & 5E-4 & 2E-3 & 1E-3  &  5E-4 \\
& Batch Size & 4 & 64 & 64  & 64\\
& Training Step/Epoch & 3000 & 60 & 2    & 5   \\
& Hidden Size & 32 & 32 & 32   & 32 \\
\midrule
\multirow{3}{*}{BitFit}
& Learning Rate & 8E-4 & 2E-3 & 1E-3  &  5E-4\\
& Batch Size & 4 & 64 & 64  &   64 \\
& Training Step/Epoch & 3000 & 60 & 2   & 5    \\
\bottomrule
\end{tabular}
}
\caption{Hyper-parameters for four training methods. We report the training steps of FewRel and the training epochs of Wiki80 and Wiki-ET.}
\label{tab:hyper-parameters-fine-tuning}
\end{table}

\subsection{General Map-tuning}

For general map-tuning, we search the dropout rate in \{0.15, 0.25, 0.35, 0.45\}. We train all the mapping networks using Adam~\cite{adam}. The learning rate is 3E-5 and the batch size is 64. We train the mapping network on the Wikipedia corpus for 5 epochs. The hyper-parameters of the best mapping network in all cases are listed in Table~\ref{tab:hyper-parameters-general}. When we evaluate RA on these datasets, we set the sequence length to 512.

\begin{table}[]
\centering
\tiny
\begin{tabular}{l|l|r|r|r|r}
\toprule
Method &  & FewRel & Wiki80 & Wiki-ET  & EntityQuestions     \\
\midrule
\multirow{2}{*}{Fine-tuning}
& Dropout & 0.25 & 0.25 & 0.25  &  0.35 \\
& Epoch & 3 & 5 & 3  & 3 \\
\midrule
\multirow{2}{*}{LoRA}
& Dropout & 0.35 & 0.25 & 0.35  & 0.25 \\
& Epoch & 5 & 5 & 4   &  5\\
\midrule
\multirow{2}{*}{Adapter}
& Dropout & 0.35 & 0.35 & 0.15  & 0.25\\
& Epoch & 5 & 5 & 5  & 5\\
\midrule
\multirow{2}{*}{BitFit}
& Dropout & 0.25 & 0.35 & 0.35  & 0.35 \\
& Epoch & 5 & 4 & 4 & 5\\
\bottomrule
\end{tabular}
\caption{Hyper-parameters for general map-tuning.}
\label{tab:hyper-parameters-general}
\end{table}

\subsection{Task-specifc Map-tuning}

We report hyper-parameters for task-specific map-tuning in Table~\ref{tab:hyper-parameters-task-specific}. We train all mapping networks using Adam with 10\% warming-up steps

Regarding the results reported in Table~\ref{tab:task-specific-map-tuning}, during task-specific map-tuning, we use dropout in the attention probabilities and all fully connected layers of the PLM. The dropout rate is 0.30, 0.20, and 0.00 for Wiki80, Wiki-ET, and EntityQuestions, respectively. Regarding the results reported in Table~\ref{tab:domain-adaptation-results}, when using training data from the source domain for task-specific map-tuning, the dropout rate is 0.35. In these cases, the training data for task-specific map-tuning are identical to those for fine-tuning the downstream models. We search the dropout rate in \{0.00, 0.15, 0.20, 0.25, 0.30, 0.35\}. When using training data from the target domain for task-specific map-tuning, we do not use dropout.

The hyper-parameters for experiments with RoBERTa are identical to those with BERT.

\begin{table}[]
\centering
\tiny
\begin{tabular}{l|r|r|r|r}
\toprule
Hyper-parameters & FewRel & Wiki80 & Wiki-ET  &   EntityQuestions   \\
\midrule
Learning Rate & 2E-5 & 4E-4 & 5E-5  &  2E-3 \\
Batch Size & 4 & 64 & 128  &  64\\
Training Step/Epoch & 3000 & 30 & 2  &  5     \\
\bottomrule
\end{tabular}
\caption{Hyper-parameters for task-specific map-tuning.}
\label{tab:hyper-parameters-task-specific}
\end{table}

\begin{table}[]
\centering
\scriptsize
\begin{tabular}{l|r|r|r|r}
\toprule
& FewRel & Wiki80 & Wiki-ET & EntityQuestions    \\
\midrule
Training Epoch & 5 & 2 & 2  & 4  \\
\bottomrule
\end{tabular}
\caption{Hyper-parameters for map-tuning on the Wikipedia corpus, after which we fine-tune BERT on downstream tasks with the mapping network plugged.}
\label{tab:hyper-parameters-fine-tuning-with-map-2}
\end{table}

\begin{table}[t]
\centering
\scriptsize
\begin{tabular}{l|r|r|r|r}
\toprule
 & FewRel & Wiki80 & Wiki-ET   & EntityQuestions\\
\midrule
Learning Rate & 2E-4 & 2E-4 & 1E-5  &  2E-4     \\
Training Epoch & 3 & 12 & 2  &  2 \\
\bottomrule
\end{tabular}
\caption{Hyper-parameters for map-tuning on downstream data, after which we fine-tune BERT on downstream tasks with the mapping network plugged.}
\label{tab:hyper-parameters-fine-tuning-with-map}
\end{table}

\subsection{Fine-tuning with the Mapping Network}

Regarding the results reported in Table~\ref{fig:fine-tuning}, the hyper-parameters for fine-tuning BERT are identical to those in Table~\ref{tab:hyper-parameters-fine-tuning}. We train all mapping networks using Adam without dropout, and the batch size is 64. For map-tuning on the Wikipedia corpus, the learning rate is 1E-5. We report other hyper-parameters for map-tuning on the Wikipedia corpus in Table~\ref{tab:hyper-parameters-fine-tuning-with-map-2}, and those for map-tuning on downstream data in Table~\ref{tab:hyper-parameters-fine-tuning-with-map}.

\subsection{Details of K-Adapter}

We use the open-source implementation of K-Adapter\footnote{\url{https://github. com/microsoft/k-adapter}}, and we only consider facAdapter (Factual Adapter). The BERT$_\textrm{base}$ layers where adapter layers plug in are $\{ 5,10 \}$. The hyper-parameters for pre-training facAdapter are identical to those reported in~\citet{K-Adapter}.

In order to plug K-Adapter into frozen downstream models in the setting of general plug-and-play injection, we tune the final fully connected layer on downstream data. We use Adam with 10\% warming-up steps, and other hyper-parameters are listed in Table~\ref{tab:hyper-parameters-k-adapter}.

\subsection{Details of Data Preprocessing}

For FewRel and Wiki80, we mark the subject and object spans by \# and \$ tokens respectively. For WikiET and EntityQuestions, we mark the entity span by \$ token.

To evaluate encoder PLMs on EntityQuestions, we append the [MASK] token to the question, and only keep the instances whose answers are in the PLM token vocabulary. We train the model to fill in the [MASK] token. It is a classification task, where all tokens in the vocabulary are choices. Only when the answer token is ranked as the top 1 result is the model considered to give a correct prediction. We further remove the instances whose entity is not in the database. Finally, we have 37800 training instances, 4693 validation instances, and 4731 test instances.

FewRel, Wiki80, and WikiET provide the annotation of entity linking, and for EntityQuestions we do entity liking by string matching.

\begin{table}[]
\centering
\tiny
\begin{tabular}{l|r|r|r|r}
\toprule
& FewRel & Wiki80 & Wiki-ET  & EntityQuestions   \\
\midrule
Learning Rate & 2E-5 & 5E-5 & 5E-5 & 5E-3 \\
Bacth Size & 4 & 32 & 64  &   64\\
Training Step/Epoch & 3000 & 15 & 2 &   20\\
\bottomrule
\end{tabular}
\caption{Hyper-parameters for tuning the final fully connected layer, during which we plug frozen K-Adapter into frozen downstream models.}
\label{tab:hyper-parameters-k-adapter}
\end{table}

\begin{table*}
\centering
\scriptsize
\begin{tabular}{l|rrrr|r|r|r}
\toprule
\multirow{2}{*}{Method} & \multicolumn{4}{c|}{FewRel 1.0} & \multirow{2}{*}{Wiki80} & \multirow{2}{*}{Wiki-ET}  & \multirow{2}{*}{EntityQuestions} \\ 
&               \multicolumn{1}{c}{5-1}     &      \multicolumn{1}{c}{5-5}     &     \multicolumn{1}{c}{10-1}     &     \multicolumn{1}{c|}{10-5}     &         &  & \\ \midrule
Fine-tuning & 1.300$\pm$0.300 & 0.800$\pm$0.436 & 2.033$\pm$0.577  & 0.533$\pm$0.231 & 0.600$\pm$0.200 & $-$0.567$\pm$0.306 & 6.967$\pm$0.850 \\
LoRA & 1.633$\pm$0.153 & 0.833$\pm$0.115 & 2.800$\pm$0.361  &  0.833$\pm$0.115  &  0.600$\pm$0.100 & 1.000$\pm$0.200 & 7.000$\pm$0.173 \\
Adapter & 1.367$\pm$0.058 & 0.733$\pm$0.115 & 2.067$\pm$0.208  &  0.833$\pm$0.153  &  0.267$\pm$0.306  &  1.100$\pm$0.529  &  6.967$\pm$0.252 \\
BitFit & 1.367$\pm$0.208  &  0.500$\pm$0.265 & 2.333$\pm$0.153  &  0.867$\pm$0.058  &  0.700$\pm$0.300 & 0.700$\pm$0.173  &  7.233$\pm$0.153 \\
 \bottomrule
\end{tabular}
\caption{The mean and standard deviation of performance improvement brought by map-tuning. We train PLMs on each downstream task with three different seeds.}
\label{tab:general-mean-var}
\end{table*}

\section{Stability of Map-tuning}

We evaluate the stability of map-tuning in general plug-and-play knowledge injection. Training the PLMs on downstream tasks with three different seeds (one of which is used in all main experiments), for each task, we have three different downstream models, into which we plug the mapping network. The mean and standard deviation of performance improvement brought by map-tuning is shown in Table~\ref{tab:general-mean-var}. From this table, we observe that map-tuning is not sensitive to downstream models overall, showing its decent stability.

\section{How Map-tuning Works with Other PLMs?}
  
In this section, we experiment map-tuning with RoBERTa~\cite{RoBERTa}, another representative PLM, on the domain transfer setting using task-specific map-tuning. The setting is identical to that in Section~\ref{sec:task-specific}. The results are shown in Table~\ref{tab:roberta}. From this table, we observe that task-specific map-tuning significantly improves the performance of the model trained on the source domain by introducing the knowledge of the target domain. Moreover, the model plugged with map-tuning is even much better than the model trained on multiple domains. It indicates that map-tuning is a universal knowledge injection method for different PLMs.

\begin{table}[t]
  \small
  \centering
  \begin{tabular}{lc|cccc}
  \toprule
  Training Data & Map & 5-1 & 5-5 & 10-1 & 10-5 \\ 
  \midrule
  \multirow{1}{*}{Target Domain}      & $-$  & 81.9 & 91.0 & 74.2 & 84.0 \\ \midrule
  \multirow{1}{*}{Multiple Domains} & $-$  & 80.9 & 92.2 & 75.4 & 87.8 \\ 
  \midrule
  \multirow{2}{*}{Source Domain}        & $-$  & 72.5 & 89.2 & 65.2 & 83.3 \\
  & \Checkmark & \textbf{91.6} & \textbf{96.6} & \textbf{88.1} & \textbf{94.5} \\
  \bottomrule
  \end{tabular}
  \caption{Results of domain adaptation using RoBERTa. We report the performance on the target domain.}
  \label{tab:roberta}
  \end{table}

\section{Empirical Analysis of MMLM}

We conduct an empirical analysis of what MMLM trains the mapping network to learn. Concretely, we split the general map-tuning corpus into a training set and a test set. During training on the training set, we plug $\mathcal{M}(\mathbf{e}_1)$ and $\mathcal{M}(\mathbf{e}_2)$ before two entity mentions $e_1$ and $e_2$ for each instance, and mask only the mention span of $e_1$. During inference on the test set, we evaluate the MMLM loss in four settings. (1) \textbf{No-Perturbation} plugs the $\mathcal{M}(\mathbf{e}_1)$ and $\mathcal{M}(\mathbf{e}_2)$, which is identical to the setting of training. (2) \textbf{Self-Perturbation} replaces $\mathcal{M}(\mathbf{\mathbf{e}_1})$ with $\mathcal{M}(\mathbf{\mathbf{e}_i})$, where $e_i$ is a random entity. (3) \textbf{Other-Perturbation} replaces $\mathcal{M}(\mathbf{\mathbf{e}_2})$ with $\mathcal{M}(\mathbf{\mathbf{e}_i})$. (4) \textbf{All-Perturbation} replaces both $\mathcal{M}(\mathbf{\mathbf{e}_1})$ and $\mathcal{M}(\mathbf{\mathbf{e}_2})$ with random ones. We also evaluate these settings with a randomly-initialized mapping network without map-tuning. For analysis, we report the result in the setting \textbf{No-Plug} where there is no plugged embedding.

\begin{table}[]
\scalebox{0.8}{
\begin{tabular}{l|r|r}
\toprule
Map-Tuning & Evaluation Setting & Loss on Test Set    \\
\midrule
$-$ & No-Plug & 7.246   \\
\midrule
\multirow{4}{*}{\Checkmark}
& No-Perturbation &   5.316 \\
& Self-Perturbation &   6.347 \\
& Other-Perturbation & 5.501       \\
& All-Perturbation & 6.613       \\
\midrule
\multirow{4}{*}{\XSolid}
& No-Perturbation & 7.179   \\
& Self-Perturbation & 7.237   \\
& Other-Perturbation & 7.268       \\
& All-Perturbation & 7.355       \\
\bottomrule
\end{tabular}
}
\caption{The MMLM loss on the test set in different evaluation settings.}
\label{tab:MMLM-analysis}
\end{table}

The result is shown in Table~\ref{tab:MMLM-analysis}. From this table, we have three observations. (1) With map-tuning, the loss in Self-Perturbation is significantly larger than that in No-Perturbation, even close to that in All-Perturbation. It proves that MMLM trains the mapping network to extract the entity information stored in the knowledge embedding so that PLMs can utilize the information. (2) The loss in Other-Perturbation is also larger than that in No-Perturbation, which indicates that the mapping network learns to extract the connections between different entities and to feed such information into PLMs. (3) Interestingly, the loss in All-Perturbation with map-tuning is smaller than that in No-Plug, and the loss in settings without map-tuning is close to the latter. The trained mapping network may be able to convert an arbitrary knowledge embedding to an embedding that can activate the PLM's own memory of some factual knowledge. In conclusion, the three mentioned abilities of mapping networks trained by MMLM enable PLMs to know new knowledge or better recall their own knowledge. Future work may improve MMLM to get stronger mapping networks.

\begin{table*}[t]
\small
\centering
\begin{tabular}{lc|rrrr|r|r|r}
\toprule
\multirow{2}{*}{Method} & \multirow{2}{*}{Map-tuning Corpus} & \multicolumn{4}{c|}{FewRel 1.0} & \multirow{2}{*}{Wiki80} & \multirow{2}{*}{Wiki-ET} & \multirow{2}{*}{EntityQuestions}\\
                       & & 5-1   & 5-5   & 10-1   & 10-5  &    &    &   \\ \midrule
Fine-tuning          & $-$           & 91.0 & 95.1 & 85.4 & 90.8 & 86.1 & 77.5  &   41.7     \\ \midrule
+ E-BERT             & $-$            & 92.3 & 95.6 & 87.6 & 91.4 & 87.8 & 79.0  &   61.3    \\
+ PELT              & $-$            & 91.2 & 95.8 & 86.1 & 91.6 & 88.2 & 79.6    &  \textbf{62.9}   \\
\midrule
\multirow{2}{*}{+ General Map} &  Wikipedia Corpus           & \textbf{93.7} & \textbf{96.2} & \textbf{89.6} & \textbf{92.4} & 88.8 & 79.9  &   \textbf{62.9}   \\ 
 & Downstream Data & 93.2 & \textbf{96.2} & 88.2 & 92.0 & \textbf{89.1} & \textbf{81.0}       &  62.0 \\
\bottomrule  
\end{tabular}
\caption{Results of knowledge injection during fine-tuning. For general map-tuning, we can use the Wikipedia corpus mentioned in the previous section or use the data of downstream tasks.}
\label{fig:fine-tuning}
\vspace{-1em}
\end{table*}

\section{Is Map-tuning Competitive in the Traditional Paradigm?}
\label{sec:fine-tuning}

It is natural to use the general mapping network in the traditional injection during fine-tuning paradigm, as the general network essentially builds an entity embedding lookup table. We freeze the parameters of the mapping network and fine-tune the PLM on downstream tasks, during which we augment model inputs with mapped knowledge representations.
Intuitively, the models learn to effectively extract information from mapped knowledge representations during fine-tuning. 
Inspired by ULMFiT~\cite{howard-ruder-2018-universal}, we also experiment on the setting where we use the task's training data as the corpus for general map-tuning. Our results are shown in Table~\ref{fig:fine-tuning}.

From this table, we have two observations: (1) map-tuning consistently outperforms E-BERT and PELT in the traditional paradigm. Considering that E-BERT and map-tuning use the same knowledge embedding, we suggest that map-tuning provides more useful knowledge representations for BERT than E-BERT. (2) General map-tuning on downstream data achieves comparable performance to that on the large-scale unsupervised corpus. It indicates that general map-tuning does not necessitate a large amount of training data for a specific task.

\section{How do We Ensure the Generality of Map-tuning?}

\begin{figure}
    \centering
    \includegraphics[width=0.9\linewidth]{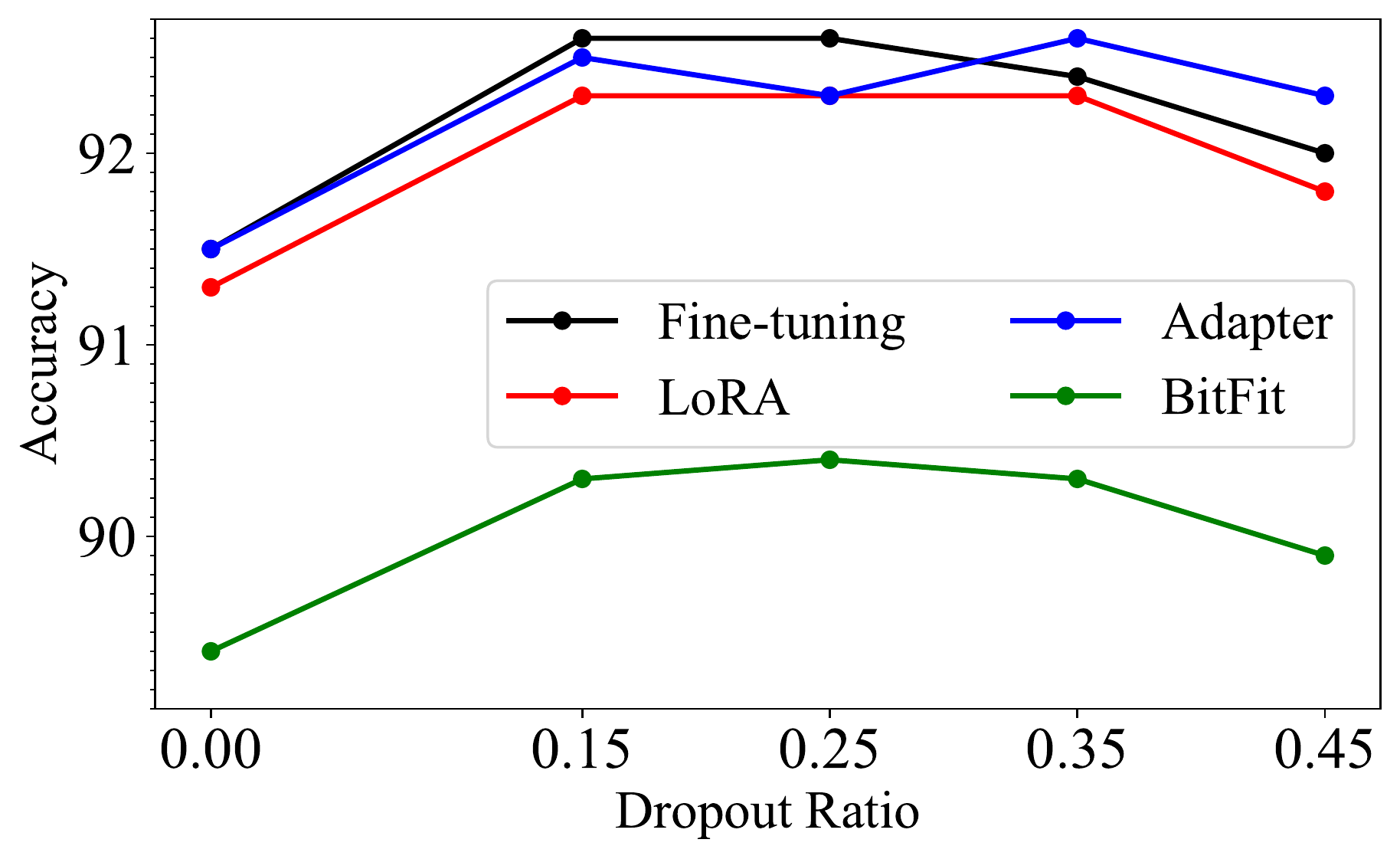}
    \caption{Effect of dropout rates on the performance of general map-tuning.}
    \label{fig:dropout}
\end{figure}

In the setting of general plug-and-play injection, we train a general mapping network based on a PLM and directly plug it into various downstream models during inference. There exists a gap between the general map-tuning procedure and the inference on downstream tasks, i.e., the PLM used for map-tuning is different from downstream models. To reduce this gap, we use dropout~\cite{Dropout} in the attention probabilities and all fully connected layers of the PLM during general map-tuning. Intuitively, dropout simulates different variants of the PLM and makes the mapping network have better generality for different downstream models trained from the PLM. We explore five different dropout rates. The results on the 5-way 1-shot of FewRel 1.0 are chosen as the representative and shown in Figure~\ref{fig:dropout}.

From this figure, we have two observations: (1) Training without dropout leads to the worst performance, which indicates that the generality of the mapping network is not good enough and downstream models can not utilize the knowledge. (2) Large dropout rates are also not optimal. Empirically, the dropout rate of 0.25 is a good choice.

\section{Numbers of Optimized Parameters}
\label{sec:parameters}

\begin{table}[t]
\centering
\small
\begin{tabular}{ccccc}
\toprule
Map-tuning & PELT & ERNIE & KEPLER & LUKE \\
\midrule
 0.1M & 123M & 114M & 123M & 274M\\
\bottomrule
\end{tabular}
\caption{Number of parameters optimized in knowledge injection. These methods are based on backbone PLMs with around 100 million parameters.}
\label{tab:parameters-optimized}
\end{table}

\begin{table*}[h]
\centering
\scriptsize
\begin{tabular}{ll|llll|l|l|l}
\toprule
\multirow{2}{*}{Method} & \multirow{2}{*}{Embedding} & \multicolumn{4}{c|}{FewRel 1.0} & \multirow{2}{*}{Wiki80} & \multirow{2}{*}{Wiki-ET}  & \multirow{2}{*}{EntityQuestions} \\ 
&            &     \multicolumn{1}{c}{5-1}     &      \multicolumn{1}{c}{5-5}     &     \multicolumn{1}{c}{10-1}     &     \multicolumn{1}{c|}{10-5}     &         &  \\ 
\midrule    \multirow{2}{*}{Fine-tuning} & $-$          &    91.0    &    95.1    &    85.4    &    90.8    &    86.1    & 77.5  & 41.7 \\
& TransE & 92.6 ($+$1.6) & 95.6 ($+$0.5) & 88.1 ($+$2.7) & 91.2 ($+$0.4) & 86.7 ($+$0.6) & 76.6 ($-$0.9)   & 49.0 ($+$7.3)  \\ 
& TransR & \textbf{93.0} ($+$\textbf{2.0}) & \textbf{95.9} ($+$\textbf{0.8}) & \textbf{88.2} ($+$\textbf{2.8}) & \textbf{92.0} ($+$\textbf{1.2}) & \textbf{86.8} ($+$\textbf{0.7}) & 77.3 ($-$0.2)   & \textbf{49.3} ($+$\textbf{7.6})  \\ \midrule
\multirow{2}{*}{LoRA}        & $-$          &    90.7    &    95.1    &    84.9    &    91.2    &    85.3    & 77.5    &   42.4    \\
& TransE & 92.3 ($+$1.6) & 96.0 ($+$0.9) & 87.4 ($+$2.5) & 91.9 ($+$0.7) & 85.8 ($+$0.5) & 78.3 ($+$0.8)  &  49.6 ($+$7.2) \\
& TransR & \textbf{92.7} ($+$\textbf{2.0}) & \textbf{96.2} ($+$\textbf{1.1}) & \textbf{87.7} ($+$\textbf{2.8}) & \textbf{92.2} ($+$\textbf{1.0}) & \textbf{86.2} ($+$\textbf{0.9}) & \textbf{78.9} ($+$\textbf{1.4})  &  \textbf{50.4} ($+$\textbf{8.0}) \\ \midrule
\multirow{2}{*}{Adapter}     & $-$          &    91.2    &    95.2    &    86.2    &    91.1    &    85.7    & 77.5   &  43.6  \\
& TransE & 92.6 ($+$1.4) & 95.8 ($+$0.6) & 88.2 ($+$2.0) & 91.8 ($+$0.7) & 85.9 ($+$0.2) & 79.2 ($+$1.7)    &  50.8 ($+$7.2)\\
& TransR & \textbf{92.9} ($+$\textbf{1.7}) & \textbf{96.0} ($+$\textbf{0.8}) & \textbf{88.4} ($+$\textbf{2.2}) & \textbf{92.3} ($+$\textbf{1.2}) & \textbf{86.4} ($+$\textbf{0.7}) & \textbf{79.5} ($+$\textbf{2.0})    &  \textbf{51.2} ($+$\textbf{7.6})\\ \midrule
\multirow{2}{*}{BitFit}      & $-$          &    89.2    &    94.8    &    83.0    &    90.0    &    82.7    & 77.1   & 41.3\\
& TransE & 90.4 ($+$1.2)  & 95.5 ($+$\textbf{0.7}) & 85.2 ($+$2.2) & 90.8 ($+$0.8) & 83.7 ($+$1.0) & 78.0 ($+$0.9)  &  \textbf{48.4} ($+$\textbf{7.1}) \\
& TransR & \textbf{91.0} ($+$\textbf{1.8}) & \textbf{95.5} ($+$\textbf{0.7}) & \textbf{85.5} ($+$\textbf{2.5}) & \textbf{91.1} ($+$\textbf{1.1}) & \textbf{83.9} ($+$\textbf{1.2}) & \textbf{78.4} ($+$\textbf{1.3})  &  48.2 ($+$6.9) \\ \bottomrule
\end{tabular}
\caption{Results of general map-tuning with different knowledge embeddings.}
\vspace{-1em}
\label{tab:transr}
\end{table*}

\begin{table}[]
\centering
\tiny
\begin{tabular}{lllll}
\toprule
           & Fine-tuning & LoRA & Adapter & BitFit \\ \midrule
-          &   35.2                           & 36.7                          & 38.1 &    35.6    \\
E-BERT     &  36.9 ($+$1.7)                   & 38.4 ($+$1.7)                 & 39.2 ($+$1.1) & 35.8 ($+$0.2)  \\
PELT       &  38.8 ($+$3.6)                   & 40.6 ($+$3.9)                 & 41.6 ($+$3.5) &  38.5 ($+$2.9)\\
RA         &  \textbf{42.7} ($+$\textbf{7.5}) & 29.0 ($-$7.7)                 & 25.0 ($-$13.1) & 17.4 ($-$18.2)  \\
K-Adapter  &   32.3 ($-$2.9)                  & 35.8 ($-$0.9)                 & 35.8 ($-$2.3) &  35.7 ($+$0.1)  \\
Map-tuning &   41.9 ($+$6.7)                  & \textbf{42.8} ($+$\textbf{6.1})     & \textbf{44.4} ($+$\textbf{6.3}) & \textbf{41.1} ($+$\textbf{5.5})      \\ \bottomrule
\end{tabular}
\caption{Performance on filtered EntityQuestions.}
\label{tab:remove_stringmatch}
\end{table}

Compared to previous knowledge injection methods, map-tuning is a parameter-efficient method. The numbers of optimized parameters for different knowledge injection methods are shown in Table~\ref{tab:parameters-optimized}. In order to introduce external knowledge, previous methods usually optimize all parameters during pre-training and fine-tuning while map-tuning only optimizes additional 0.1\% of parameters and freezes the original model, which makes it flexible to use mapping networks for different inputs with the same models.

\section{Performance with Relation Information}

Inspired by TransR~\cite{lin2015learning}, which proposes to incorporate relation information into the entity representations, we also experiment with the relation-aware entity representations in map-tuning. Based on the TransE embeddings used in the previous experiments, we calculate the average of the relation embeddings of a certain entity and concatenate it with the original entity embedding. This method can be viewed as a simplified version of TransR. The results are shown in Table~\ref{tab:transr}. From this table, we observe that TransR improves the performance of map-tuning in most cases, which indicates that relation information is also useful for map-tuning.

\section{Performance on EntityQuestions}

We report the performance on filtered EntityQuestions in Table~\ref{tab:remove_stringmatch}.

\end{document}